\documentclass{article}
\usepackage{fullpage}
\usepackage{times}
\usepackage{graphicx} 
\usepackage{subfig}

\graphicspath{{./figs/}}

\usepackage{algorithm}
\usepackage{algorithmic}

\usepackage{hyperref}
\usepackage{authblk}

\title{GLSR-VAE: Geodesic Latent Space Regularization for Variational
  AutoEncoder Architectures}
\newcommand{\dd}{\mathrm{d}}

\date{}
\begin{document}








\author[1, 2]{Ga\"etan Hadjeres}
\author[3, 4]{Frank Nielsen}
\author[2]{Fran\c cois Pachet}

 \affil[1]{LIP6, Universit\'e Pierre et Marie Curie}
 \affil[2]{Sony CSL, Paris}
 \affil[3]{\'Ecole Polytechnique, Palaiseau, France}
 \affil[4]{Sony CSL, Tokyo}







\maketitle

\begin{abstract}
  VAEs (Variational AutoEncoders) have proved to be powerful in the context of density
  modeling and have been used in a variety of contexts for creative purposes. In
  many settings, the data we model possesses continuous attributes
  that we would like to take into account at generation time.
  
  We propose in this paper GLSR-VAE, a \emph{Geodesic Latent Space
    Regularization for the Variational AutoEncoder} architecture and its
  generalizations which allows a fine control on the embedding of the data into
  the latent space. When augmenting the VAE loss with this regularization,
  changes in the learned latent space reflects changes of the attributes of the
  data. This deeper understanding of the VAE latent space structure offers the
  possibility to modulate the attributes of the generated data in a continuous
  way. We demonstrate its efficiency on a monophonic music generation task where
  we manage to generate variations of discrete sequences in an intended and playful way.
\end{abstract}

\section{Introduction}
\label{sec:introduction}
Autoencoders \cite{bengio2009learning} are useful for learning to encode
observable data into a latent space of smaller dimensionality and thus
perform dimensionality reduction (manifold learning). However, the
latent variable space often lacks of structure \cite{vincent2010stacked} and it
is impossible, by construction, to sample from the data distribution.  The
Variational Autoencoder (VAE) framework \cite{2013arXiv1312.6114K} addresses these
two issues by introducing a regularization on the latent space together with an
adapted training procedure. This allows to train complex generative models with
latent variables while providing a way to sample from the learned data
distribution, which makes it useful for unsupervised density modeling.

\begin{figure*}[h]
  \centering \subfloat[VAE encoding of the MNIST dataset in a latent variable
  space $\mathcal{Z} := \mathbf{R}^2$. Each point corresponds to a dataset point
  and colors. Reproduced from \cite{makhzani2015adversarial}.]{
    \includegraphics[height=5cm]{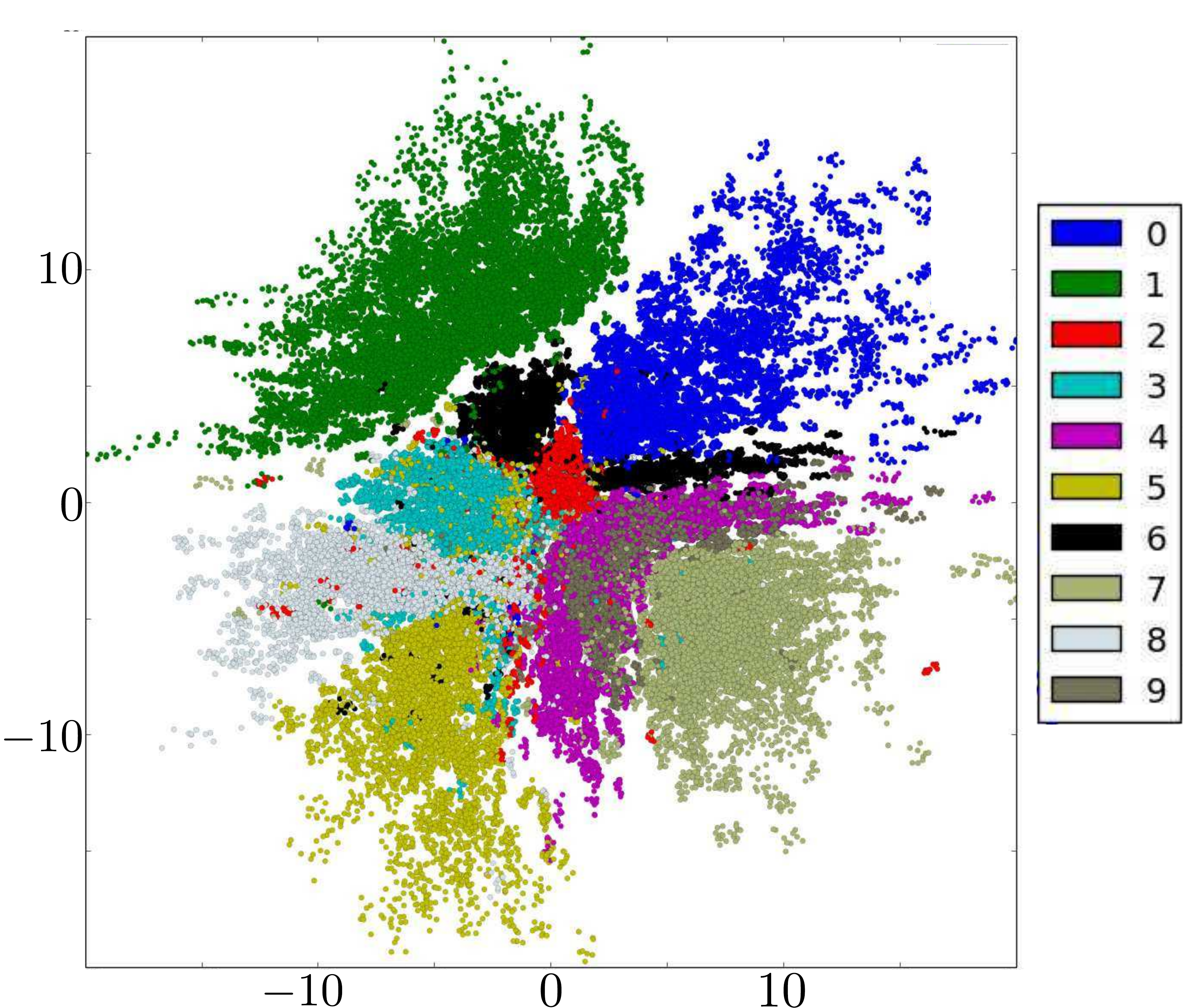}

    \label{fig:VAEmnist}

  } \quad \subfloat[Visualization of the learned data manifold. Reproduced from
  \cite{2013arXiv1312.6114K}.]{
    \includegraphics[height=5cm]{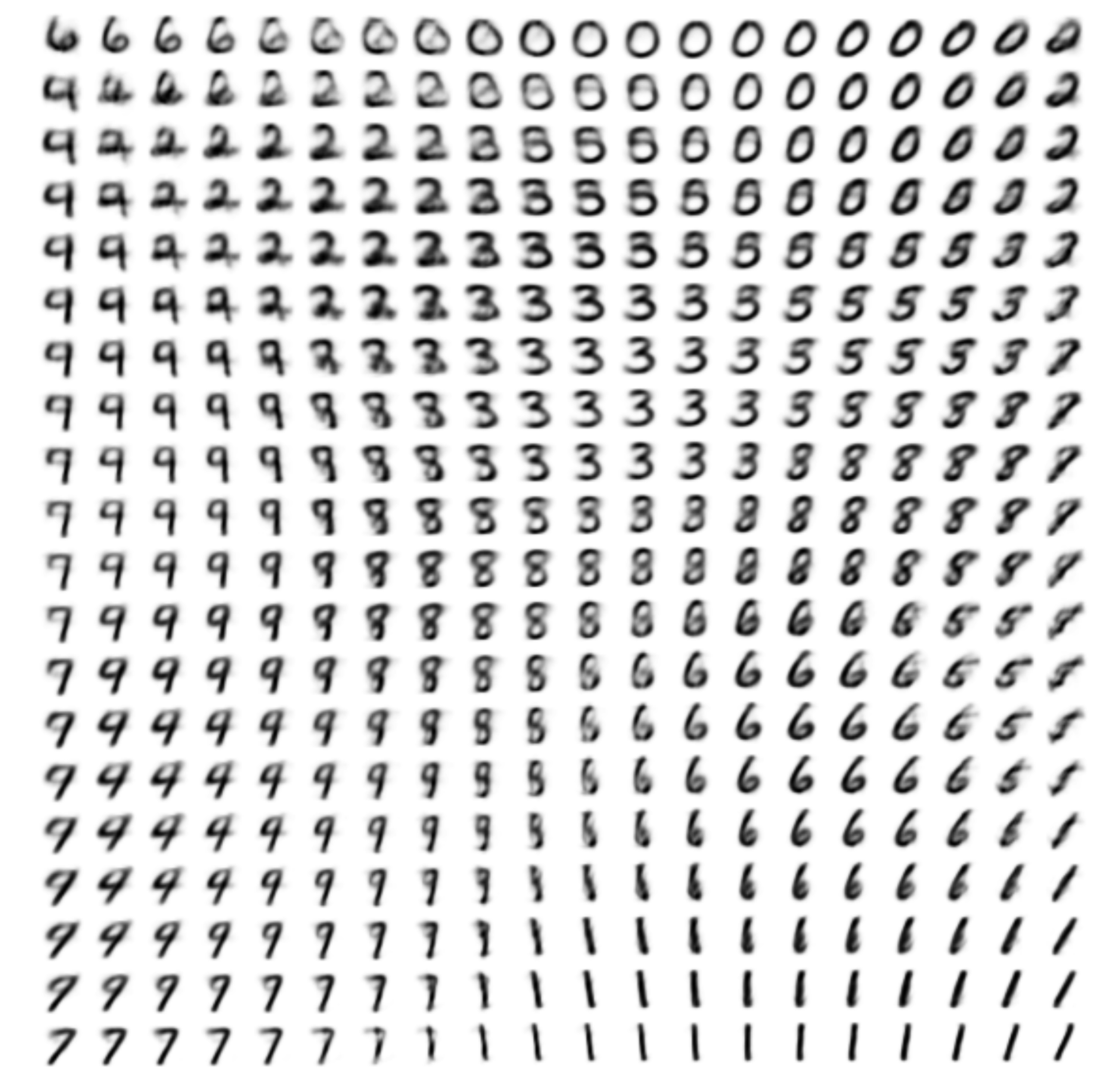}

\label{fig:MNISTmanifold}
}
\label{fig:mnist}
\caption{Two visualizations of the latent space of a VAE trained on the MNIST
  dataset.}
\end{figure*}

Once trained, a VAE provides a decoding function, i.e. a mapping from a
low-dimensional latent space to the observation space which defines what is
usually called a \emph{data manifold} (see Fig.~\ref{fig:MNISTmanifold}).  It is
interesting to see that, even if the observation space is discrete, the latent
variable space is continuous which allows one to define \emph{continuous paths}
in the observation space i.e. images of continuous paths in the latent variable
space.  This interpolation scheme has been successfully applied to image
generation \cite{2015arXiv150204623G,makhzani2015adversarial} or text
generation \cite{DBLP:journals/corr/BowmanVVDJB15}. However, any continuous path
in the latent space
can produce an interpolation in the observation space and there is no way to
prefer one over another \emph{a priori}; thus the straight line between two points will not necessary
produce the ``best'' interpolation.

When dealing with data that contains more information, such as labels or
interpretive quantities, it is interesting to see if and how this information
has been encoded into the latent space (see
Fig.~\ref{fig:VAEmnist}). Understanding the latent space structure can be of
great use for the generation of new content as it can provide a way to
manipulate high-level concepts in a creative way.

One way to control the generating process on annotated data is by conditioning
the VAE model, resulting in the Conditional Variational AutoEncoder (CVAE)
architectures \cite{NIPS2015_5775, 2015arXiv151200570Y}. These models can
be used to generate images with specific attributes but also allow to generate
interpolation between images by changing only a given attribute. In these
approaches, the CVAE latent spaces do not contain the high-level information but
the randomness of the produced images for fixed attributes.

Another approach, where is no decoupling between the attributes and the latent
space representation, consists in finding \emph{attribute vectors} which are
vectors in the latent space that could encode a given high-level feature or
concept.  Translating an encoded input vector by an attribute vector before
decoding it should ideally add an additional attribute to the data. This has
been successfully used for image generation with VAEs (coupled with an
adversarially learned similarity metric) in \cite{larsen2015autoencoding} and
even directly on the high-level feature representation spaces of a classifier
\cite{upchurch2016deep}. However, these approaches rely on finding \emph{a
  posteriori} interpretations of the learned latent space and there is no
theoretical reason that simple vector arithmetic has a particular significance
in this setting. Indeed, in the original VAE article \cite{2013arXiv1312.6114K}
the MNIST manifold (reproduced in Fig.~\ref{fig:MNISTmanifold}) has been
obtained by transforming a linearly spaced coordinate grid on the unit square
through the inverse CDF of the normal distribution in order to obtain
``equally-probable'' spaces between each decoded image. This advocates for the
fact that the latent space coordinates of the data manifold need not be
\emph{perceived} as geodesic normal coordinates. That is, decoding a  straight line drawn
in the latent
space does not give rise to elements whose attributes vary uniformly.

In this work, we focus on fixing \emph{a priori} the geometry of the latent
space when the dataset elements possess continuous (or discrete and \emph{ordered})
attributes. By introducing a \emph{Geodesic Latent Space Regularization} (GLSR), we
show that it is possible to relate variations in the latent space to variations
of the attributes of the decoded elements.

Augmenting the VAE training procedure with a regularizing term has been recently
explored in \cite{lamb2016discriminative} in the context of image generation
where the introduction of a \emph{discriminative regularization} is aimed at
improving the visual quality of the samples using a pre-trained classifier. Our
approach differs from the one above in the fact that the GLSR favors latent
space representations with fixed attribute \emph{directions} and focuses more on
the latent space structure.

We show that adding this regularization grants the latent space a meaningful
interpretation while retaining the possibility to sample from the learned data
distribution. We demonstrate our claim with experiments on musical data.  Our
experiments suggest that this regularization also helps to learn latent
variable spaces with little correlation between regularized and non regularized
dimensions.  Adding the possibility to \emph{gradually} alter a generated sample
according to some user-defined criteria can be of great use in many generative
tasks. Since decoding is fast, we believe that this technique can be used for
interactive and creative purposes in many interesting and novel ways.




\section{Regularized Variational Autoencoders}

\subsection{Background on Variational Autoencoders}
\label{sec:rappelvae}

We define a \emph{Variational AutoEncoder} (VAE) as a deep generative model
(like Generative Adversarial Networks (GANs) \cite{2017arXiv170600550H}) for
observations $x \in \mathcal{X}$ that depends on latent variables
$z \in \mathcal{Z}$ by writing the joint distribution $p_\theta(x, z)$ as
\[ p_\theta(x, z) = p(z) p_\theta(x | z),\] where $p(z)$ is a \emph{prior}
distribution over $z$ and $p_\theta(x | z)$ a \emph{conditional distribution}
parame\-trized by a neural network NN$(\theta)$. Given a i.i.d. dataset
$X = \{x^1, \dots, x^N\}$ of elements in $\mathcal{X}$, we seek the parameter
$\theta$ maximizing the dataset likelihood
\begin{equation}
  \label{eq:likelihood}
  \log p_\theta(X) = \sum_{i = 1}^N \log p_\theta(x^i).
\end{equation}
However, the marginal probability
\[p_\theta(x) = \int p(x, z) \dd z\]
and the posterior probability
\[p_\theta(z | x) = \frac{p(x, z)}{p(x)} = \frac{p(x, z)}{\int p(x, z) \dd z} \]
are generally both computationally intractable which makes maximum likelihood
estimation unfeasible. The solution proposed in \cite{2013arXiv1312.6114K}
consists in preforming Variational Inference (VI) by introducing a parametric
variational distribution $q_\phi(z | x)$ to approximate the model's posterior
distribution and lower-bound the marginal log-likelihood of an observation $x$;
this results in:
\begin{equation}
  \label{eq:elbo}
  \log p(x) \geq \mathbf{E}_{q_\phi(z|x)} \left[ \log p(x |z)\right] - D_{KL}(q(z|x) || p(z)) := \mathcal{L}(x; \theta, \phi),
\end{equation}
where $D_{KL}$ denotes the Kullback-Leibler divergence \cite{cover2012elements}. 

Training is performed by maximizing the \emph{Evidence Lower BOund} (ELBO) of
the dataset
\begin{equation}
  \label{eq:elbodataset}
\mathcal{L}(\theta, \phi) := \sum_{i = 1}^N \mathcal{L}(x^i; \theta, \phi)
\end{equation}
by jointly optimizing over the parameters $\theta$ and $\phi$.  Depending on the
choice of the prior $p(z)$ and of the variational approximation $q_\phi(z|x)$,
the Kullback-Leibler divergence $D_{KL}(q(z|x) || p(z))$ can either be computed
analytically or approximated with Monte Carlo integration.

Eq.~(\ref{eq:elbo}) can be understood as an autoencoder with stochastic units
(first term of $\mathcal{L}(x; \theta, \phi)$) together with a regularization
term given by the Kullback-Leibler divergence between the approximation of the
posterior and the prior. In this analogy, the distribution $q_\phi(z | x)$ plays
the role of the \emph{encoder} network while $p_\theta(x | z)$ stands for the
\emph{decoder} network.

\subsection{Geodesic Latent Space Regularization (GLSR)}
\label{sec:reg}
We now suppose that we have access to additional information about the
observation space $\mathcal{X}$, namely that it possesses ordered quantities of
interest that we want to take into account in our modeling process. These
quantities of interest are given as $K$ independent differentiable real \emph{attribute
  functions} $\{g_k\}$ on $\mathcal{X}$, with $K$ less than the dimension of the
latent space.

In order to better understand and visualize what a VAE has learned, it can be
interesting to see how the expectations of the attribute functions
\begin{equation}
  \label{eq:expectation-gk}
G_k: z \mapsto \mathbf{E}_{p_\theta(x | z)}[g_k(x)] 
\end{equation}
behave as functions from  $\mathcal{Z}$ to $\mathbf{R}$. In the Information
Geometry (IG) literature \cite{amari2007methods,amari2016information}, the $G_k$ functions are called the \emph{moment
  parameters} of the statistics $g_k$.

Contrary to other approaches which try to find \emph{attribute vectors} or
\emph{attribute directions} \emph{a posteriori}, we propose to impose the directions of
interest in the latent space by linking changes in the latent space
$\mathcal{Z}$ to changes of the $G_k$ functions at training time. Indeed,
linking changes of $G_k$ (that have meanings in applications) to changes of the
latent variable $z$ is a key point for steering (interactively) generation.

This can be enforced by adding a regularization term over
$z = (z_1, \dots, z_{\textrm{dim}\mathcal{Z}})$ to the ELBO
$\mathcal{L(\theta, \phi)}$ of Eq.~(\ref{eq:elbodataset}). We define the
\emph{Geodesic Latent Space Regularization for the Variational Auto-Encoder} (GLSR-VAE) by
\begin{equation}
  \label{eq:glsr}
\mathcal{R}_{\textrm{geo}}(z; \{g_k\}, \theta) := \sum_{k=1}^K \mathcal{R}_k(z; \theta)
\end{equation}
where
\begin{equation}
  \label{eq:glsrk}
  \mathcal{R}_k(z; \theta) = \log 
  r_k \left(\frac{\partial G_k}{\partial z_k}(z)\right).
\end{equation}

The distributions 
$r_k$ over the values of the partial derivatives of $G_k$ is chosen so
that 
$\mathbf{E}_{u}[r_k(u)] > 0$, and preferably peaked around its mean
value (small variance). Their choice is discussed in appendix~\ref{sec:regpar}.

Ideally (in the case where the distributions $r_k$ are given by Direct delta
functions with strictly positive means), this regularization forces infinitesimal changes
$\dd z_k$ of the variable $z$ to be proportional (with a positive factor) to
infinitesimal changes of the functions $G_k$ (Eq.~\ref{eq:expectation-gk}). In
this case, for $z_{K+1}, \dots, z_{\textrm{dim}\mathcal{Z}} \in \mathcal{Z}$
fixed, the mapping
\begin{equation}
  \label{eq:manifoldgk}
  (z_1, \dots, z_K) \mapsto (G_1(z), \dots, G_K(z)) \in \mathbf{R}^K,
\end{equation}
where $z =   (z_1, \dots, z_K, z_{K+1},  \dots, z_{\textrm{dim}\mathcal{Z}})$
defines a Euclidean manifold in which geodesics are given by all straight lines.

To summarize, we are maximizing the following regularized ELBO:
\begin{equation}
  \label{eq:elboreg}
\mathcal{L}_\textrm{reg}(x; \theta, \phi) := \mathbf{E}_{q_\phi(z|x)} \left[ \log p(x |z) + \mathcal{R}_{\textrm{geo}}(z; \{g_k\}, \theta)\right] - D_{KL}(q(z|x) || p(z)).
\end{equation}
Note that we are taking the expectation of $\mathcal{R_\textrm{geo}}$ with
respect to the variational distribution $q_\phi(z | x)$.

Eq.~(\ref{eq:elboreg}) can be maximized with stochastic gradient ascent using
Monte-Carlo estimates of the intractable estimations. We also use the
reparametrization trick \cite{2013arXiv1312.6114K} on stochastic variables to
obtain low-variance gradients.


\section{Experiments}
 \label{sec:exp}
 In this section, we report experiments on training a VAE on the task of
 modeling the distribution of \emph{chorale melodies in the style of J.S. Bach}
 with a geodesic latent space regularization. Learning good latent
 representations for discrete sequences is known to be a challenging problem with
 specific issues (compared to the continuous case) as pinpointed in
 \cite{2017arXiv170604223J}.  Sect.~\ref{sec:vae} describes how we used the VAE
 framework in the context of sequence generation, Sect.~\ref{sec:data} exposes
 the dataset we considered and Sect.~\ref{sec:res} presents experimental results
 on the influence of the geodesic latent space regularization tailored for a
 musical application. A more detailed
 account on our implementation is deferred to appendix~\ref{sec:details}.

\subsection{VAEs for Sequence Generation}
\label{sec:vae}
We focus in this paper on the generation of discrete sequences of a given length
using VAEs. Contrary to recent approaches
\cite{FabiusVRAE,ChungKDGCB15,fraccaro2016sequential}, we do not use recurrent
latent variable models but encode each entire sequence in a single latent
variable.

In this specific case, each sequence $x = (x_1, \dots, x_T) \in \mathcal{X}$ is
composed of $T$ time steps and has its elements in $[A]$, where $A$ is the number
of possible tokens while the variable $z$ is a vector in $\mathcal{Z}$.

We choose the prior $p(z)$ to be a standard Gaussian distribution with zero mean and unit
variance.

The approximated posterior or encoder $q_\phi(z |x)$ is modeled using a normal
distribution $\mathcal{N}(\mu(x), \sigma^2(x))$ where the functions $\mu$ and
$\sigma^2$ are implemented by Recurrent Neural Networks (RNNs)
\cite{Goodfellow-et-al-2016}.

When modeling the  conditional distribution $p_\theta(x | z)$ on sequences from
$\mathcal{X}$, we suppose that all variables $x_i$ are independent, which means
that we have the following factorization:
\begin{equation}
  \label{eq:probarnn}
  p_\theta(x | z) := \prod_{i=1}^T p_\theta^i(x_i | z).
\end{equation}

In order to take into account the sequential aspect of the data and to make our
model size independent of the sequence length $T$, we implement
$p_\theta(x | z)$ using a RNN.  The particularity of our implementation is that
the latent variable $z$ is only passed as an input of the RNN decoder on the
first time step. To enforce this, we introduce a binary mask $m \in \{0, 1\}^T$
such that $m_1 = 1$ and $m_i=0$ for $i>1$ and finally write
\begin{equation}
  \label{eq:probarnnmasked}
  p_\theta(x | z) := \prod_{i=1}^T p_\theta^i(x_i | m_i * z, m_{<i}),
\end{equation}
where the multiplication is a scalar multiplication and where
$m_{<i} := \{m_1, \dots, m_{i-1}\}$ for $i > 1$ and is $\emptyset$ for $i=1$. In practice,
this is implemented using one RNN cell which takes as input $m_i * z, m_{i}$ and
the previous hidden state $h_{i-1}$. The RNN takes also the binary mask itself
as an input so that our model differentiates the case $z = 0$ from the case
where no latent variable is given.

The decoder $p_\theta(x | z)$ returns in fact probabilities over
$\mathcal{X}$. In order to obtain a sequence in $\mathcal{X}$ we have typically
two strategies which are: taking the maximum a posteriori (MAP) sequence
\begin{equation}
  \label{eq:argmax}
  x = \textrm{argmax}_{x' \in \mathcal{X}} p_\theta(x' | z)
\end{equation}
or sampling each variable independently (because of
Eq.~(\ref{eq:probarnnmasked})). These two strategies give rise to mappings from
$\mathcal{Z}$ to $\mathcal{X}$ which are either deterministic (in argmax
sampling strategy case) or stochastic. The mapping
\begin{equation}
  \label{eq:mapping}
z \mapsto \textrm{argmax}_{x' \in \mathcal{X}} p_\theta(x' | z)
\end{equation}
is usually thought of
defining the data manifold learned by a VAE.

Our approach is different from the one proposed in
\cite{chen2016variational} since the latent variable $z$ is only passed on the
first time step of the decoder RNN and the variables are independent. We believe
that in doing so, we ``weaken the decoder'' as it is
recommended in \cite{chen2016variational} and force the decoder to use
information from latent
variable $z$.

 We discuss more precisely the parametrization we used for the conditional
 distribution $p(x|z)$ and the approximated posterior $q(z|x)$ in
 appendix \ref{sec:details}.

\subsection{Data Preprocessing}
\label{sec:data}
We extracted all monophonic soprano parts from the J.S. Bach chorales dataset as
given in the music21 \cite{cuthbert2010music21} Python package. We chose to
discretize time with sixteenth notes and used the \emph{real name} of notes as
an encoding. Following \cite{2016arXiv161201010H}, we add an extra symbol which
encodes that a note is held and not replayed. Every chorale melody is then
transposed in all possible keys provided the transposition lies within the
original voice ranges.  Our dataset is composed of all contiguous subsequences
of length $\Delta t=32$ and we use a latent variable space with $12$ dimensions.
Our observation space is thus composed of sequences
$x = (x_1, \dots, x_{32}) \in \mathcal{X}$ where each element of the sequence
$x_i$ can be chosen between $A = 53$ different tokens.

\subsection{Experimental Results}
\label{sec:res}
We choose to regularize one dimension by using as a function $g(x) := g_1(x)$
the number of played notes in the sequence $x$ (it is an integer which is
explicitly given by the representation we use).


\subsubsection{Structure of the latent space}
\label{sec:struct}
Adding this regularization directly influences how the
embedding into the latent space is performed by the VAE. We experimentally check that
an increase $\Delta z_1$ in the first coordinate  of the latent space variable
$z = (z_1, \dots, z_{\textrm{dim}\mathcal{Z}})$ leads to an increase of
\begin{equation}
  \label{eq:gargmax}  
g_{\mathcal{Z}} := z \mapsto g(\textrm{argmax}(p_\theta(x|z))).
\end{equation}
The (non-differentiable) function (Eq.~\ref{eq:gargmax}) is in fact the real
quantity of interest, even if it is the the differentiable function $G_1$
(Eq.~\ref{eq:expectation-gk}) which is involved in the geodesic latent space
regularization (Eq.~\ref{eq:glsrk}). In order to visualize the high-dimensional
function $g_{\mathcal{Z}}$, we plot it on a 2-D plane containing the regularized
dimension. In the remaining of this article, we always consider the plane
\begin{equation}
  \label{eq:plane}
P_{z_1, z_2} = \left\{(z_1, z_2, 0, \dots, 0), z_1, z_2 \in \mathbf{R} \right\}
\end{equation}

\begin{figure}[]
\centering
  \subfloat[Without geodesic latent space regularization]{
 \includegraphics[scale=0.5]{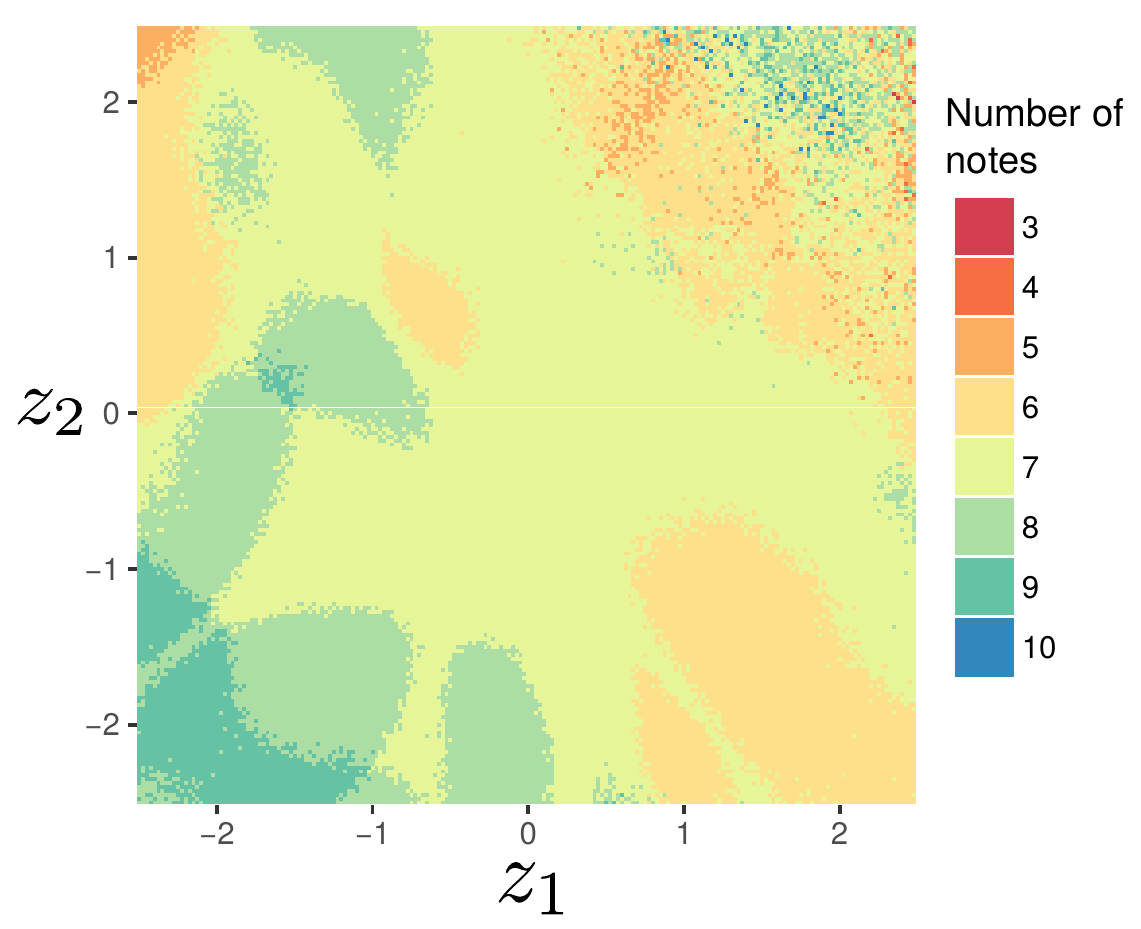}
\label{fig:numNotesNoReg}
  }
  \subfloat[With geodesic latent space regularization]{
    \includegraphics[scale=0.5]{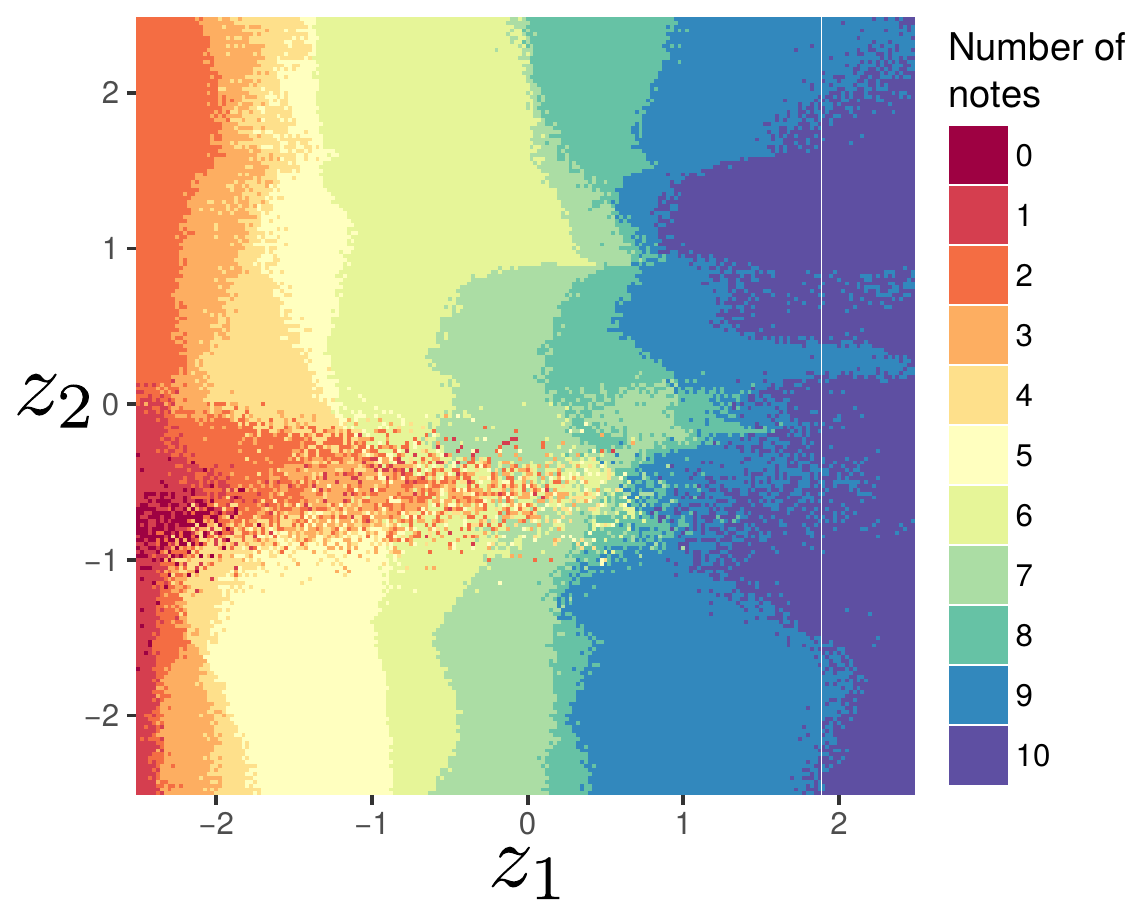}
    \label{fig:numNotesReg}
}
\caption{Plot of a 2-D plane in the latent variable space $\mathcal{Z}$. The
  x-axis corresponds to the regularized dimension.}
\label{fig:numNotes}
\end{figure}

Fig.~\ref{fig:numNotes} shows plots of the $g_{\mathcal{Z}}$ function restricted
to the 2-D plane $P_{z_1, z_2}$.
The case where
no geodesic latent space regularization is applied is visible in
Fig.~\ref{fig:numNotesNoReg} while the case where the regularization is applied
on one latent space dimension is shown in Fig.~\ref{fig:numNotesReg}. There is a
clear distinction between both cases: when the regularization is applied, the
function $g_{\mathcal{Z}}$ is an increasing function on each horizontal line
while there is no predictable pattern or structure in the non-regularized case.


\begin{figure}[]
\centering
  \subfloat[Without geodesic latent space regularization]{
 \includegraphics[scale=0.32, clip=true, trim=0 0 10 0]{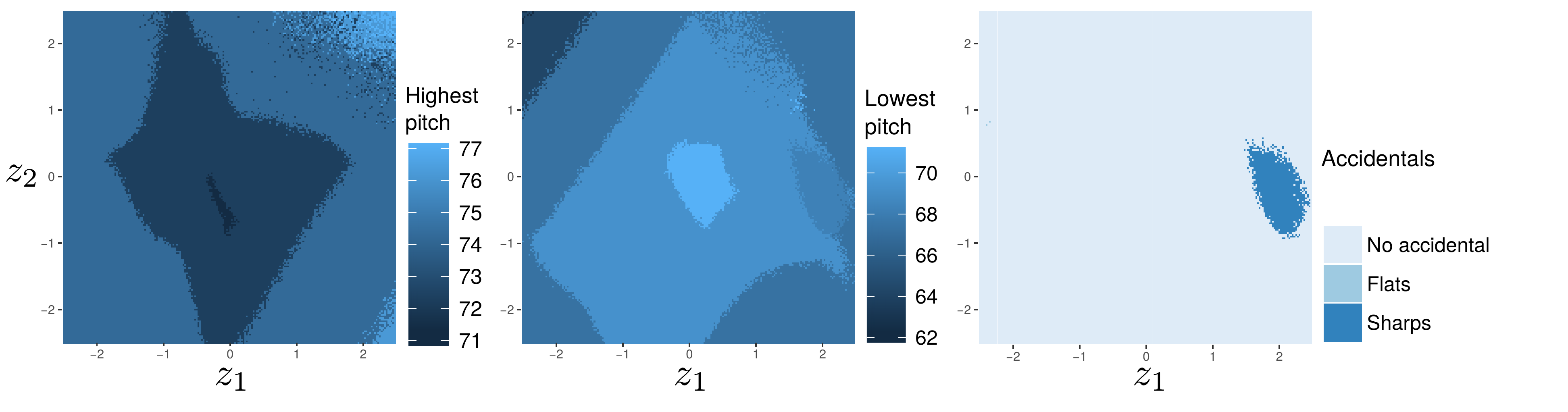}
\label{fig:multiplotNoReg}
}

  \subfloat[With geodesic latent space regularization]{
    \includegraphics[scale=0.32, clip=true, trim=10 0 0 0]{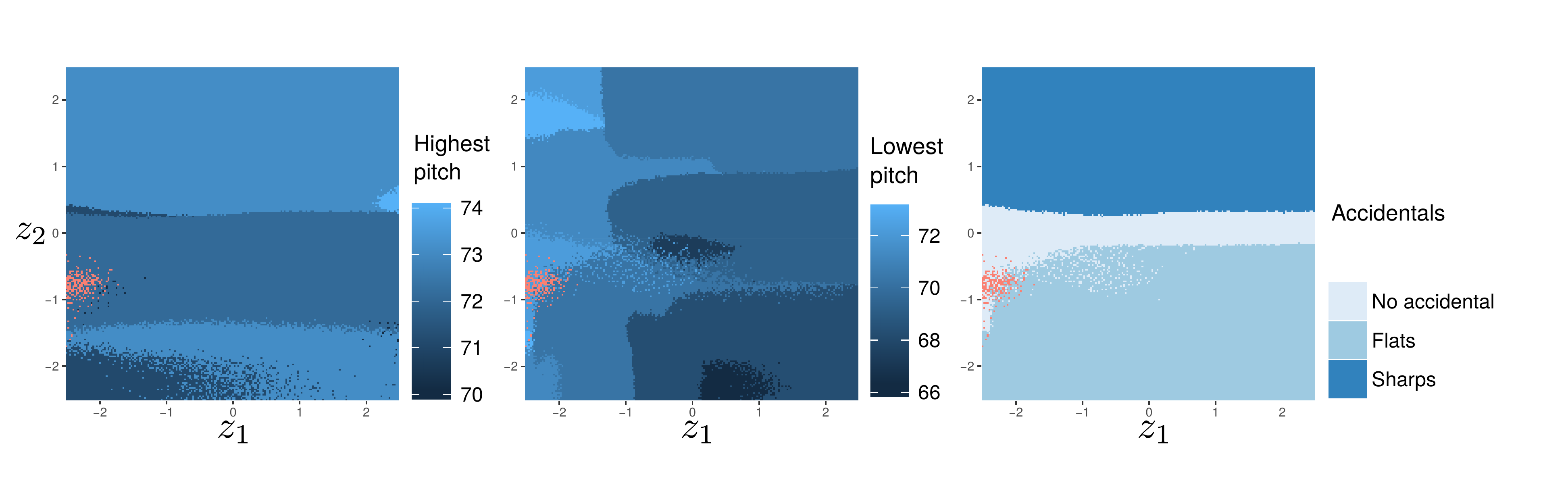}
    \label{fig:multiplotReg}
}
\caption{Plot of a 2-D plane in the latent variable space $\mathcal{Z}$. The
  x-axis corresponds to the regularized dimension. Different non-regularized
  quantities of the decoded sequences are displayed: the highest pitch, the
  lowest pitch  and if the sequence contains no accidental,
  sharps or flats.}
\label{fig:multiplot}
\end{figure}

In order to see if the geodesic latent space regularization has only effects on
the regularized quantity (given by $g_\mathcal{Z}$) or also affects other (non
regularized) attribute functions, we plot as in Fig.~\ref{fig:numNotes} these
attribute functions (considered as real functions on $\mathcal{Z}$ as in
Eq.~(\ref{eq:gargmax})).  Figure~\ref{fig:multiplot} show plots of different
attribute functions such as the highest and lowest MIDI pitch of the
sequence and the presence of sharps or flats.  We remark that adding the latent
space regularization tends to decorrelate the regularized quantities from the
non-regularized ones.

\subsubsection{Generating Variations by moving in the latent space}
\label{sec:variations}
Reducing correlations between features so that each feature best accounts for
only one high-level attribute is often a desired property
\cite{cogswell2015reducing} since it can lead to better generalization and
non-redundant representations. This kind of ``orthogonal features'' is in
particular highly suitable for interactive music generation. Indeed, from a
musical point of view, it is interesting to be able to generate variations of a
given sequence with more notes for instance while the other attributes of the
sequence remain unchanged.

\begin{figure}[]
  \centering
    \includegraphics[scale=0.63]{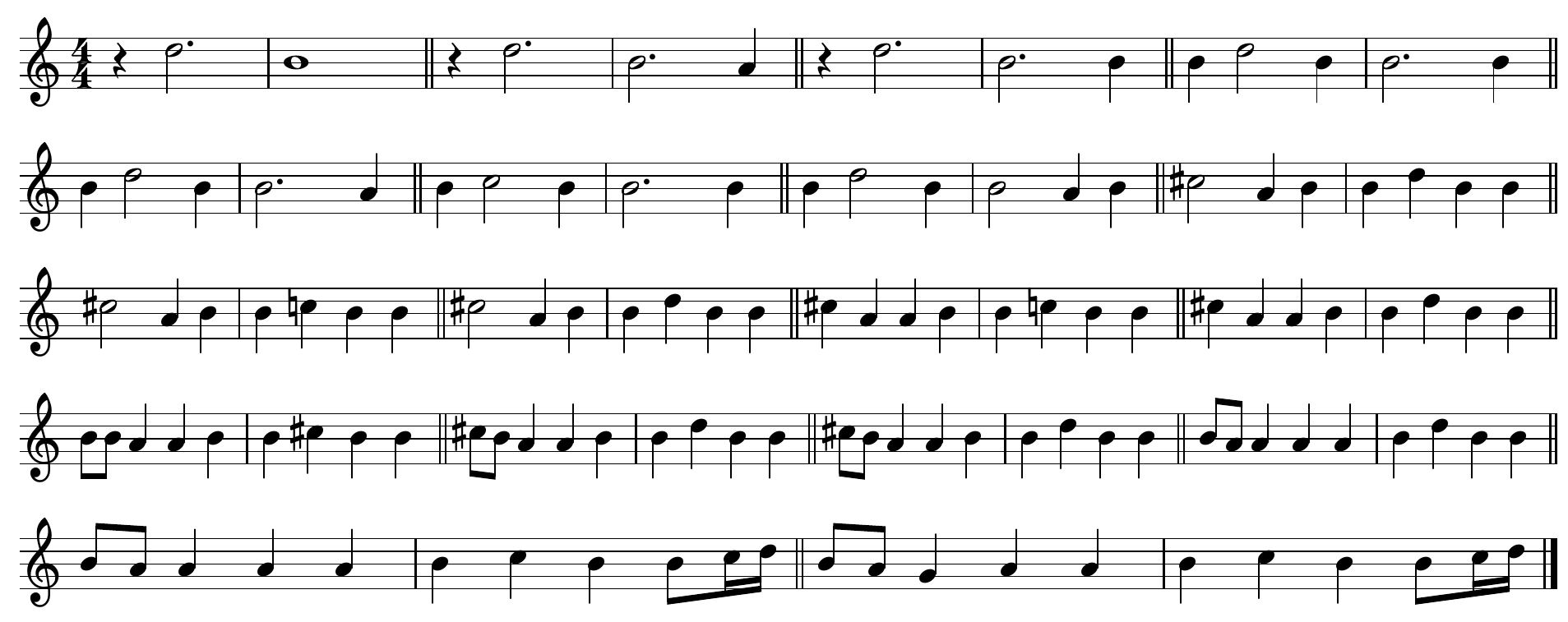}
    \caption{Image of straight line in the data manifold obtained by
      starting from a random $z$ and then only increasing its first
      (regularized) coordinate $z_1$. The argmax decoding procedure
      (Eq.~\ref{eq:argmax}) was used. All generated sequences are two-bar long
      and separated by double bar lines. This generates variations of the initial
      motif by adding more notes.}
  \label{fig:increasingNumNotes}
\end{figure}

The problem of sampling sequences with a fixed number of notes with
the correct data distribution has been, for example, addressed in
\cite{PapadopoulosRP16} in the context of sequence generation with Markov
Chains. In our present case, we have the possibility to progressively add notes
to an existing sequence by simply moving with equal steps in the
regularized dimension. We show in Fig.~\ref{fig:increasingNumNotes}
how moving only in the regularized dimension of the latent space gives rise to
variations of an initial starting sequence in an \emph{intended} way. 

\subsubsection{Effect on the aggregated distribution and validation accuracy}
\label{sec:aggdist}
A natural question which arises is: does adding a geodesic regularization on the
latent space deteriorates the effect of the Kullback-Leibler regularization or
the reconstruction accuracy?
The possibility to sample
from the data distribution by simply drawing a latent variable $z \sim p(z)$
from the prior distribution and then drawing $x \sim p_\theta(x | z)$ from the
conditional distribution indeed constitutes one of the great advantage of the VAE
architecture.

\begin{figure}[]
  \centering
    \includegraphics[scale=0.5]{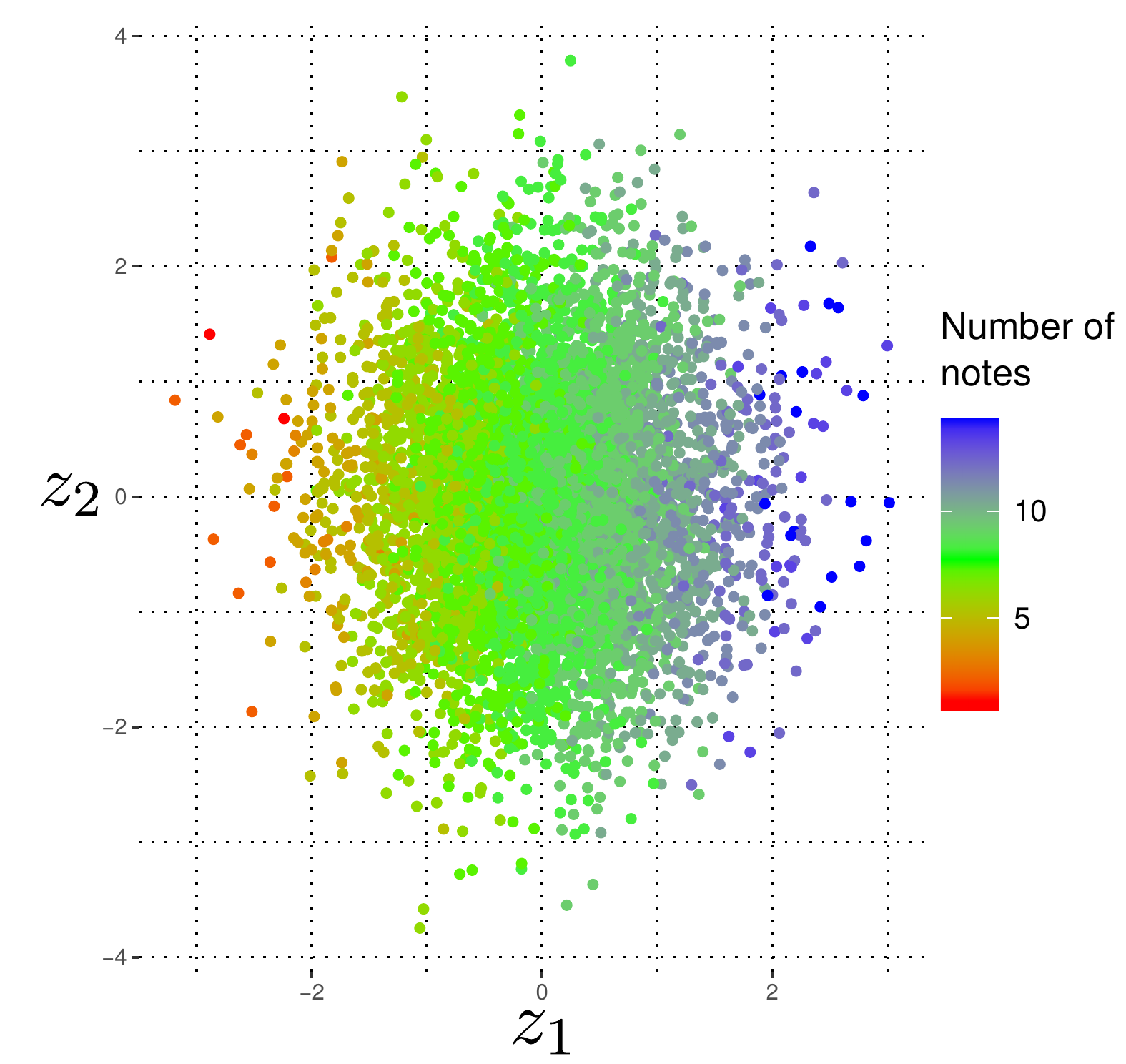}
    \caption{Plot of the aggregated distribution projected on a 2-D plane in the
      latent space which contains the regularized dimension as its x-axis. Each
      point is attributed a color depending on the number of notes contained in
      the decoded sequence.}
  \label{fig:agg}
\end{figure}

We check this by looking at the \emph{aggregated distribution} defined by
\begin{equation}
  \label{eq:agg}
q_\phi(z) := \int_x q_\phi(z |x) p_d(x) \dd x,  
\end{equation}
where $p_d(x)$ denotes the data distribution. In an ideal setting, where
$q_\phi(z|x)$ perfectly matches the posterior $p_\theta(z|x)$, the aggregated
distribution $q_\phi(z)$ should match the prior $p(z)$. We experimentally
verify this by plotting the aggregated distribution projected on a 2-D plane in
Fig.~\ref{fig:agg}. By assigning colors depending on the regularized quantity,
we notice that even if the global aggregated distribution is normally
distributed and approach the prior, the aggregated distribution of each cluster
of sequences (clustered depending on the number of notes they contain) is not,
and depends on the the regularized dimension.

We report a slight drop (1\%) of the reconstruction accuracy when adding the
geodesic latent space regularization. The fact that adding a regularization term
reduces the reconstruction accuracy has also been noted in
\cite{lamb2016discriminative} where they nonetheless report a better visual
quality for their regularized model.

The geodesic latent space regularization thus permits to obtain more meaningful
posterior distributions while maintaining the possibility to sample using the
prior distribution at the price of a small drop in the reconstruction accuracy.
We believe that devising adaptive geodesic latent space regularizations could be
a way to prevent this slight deterioration in the model's performance and
provide us with the best of both worlds.  Having the possibility to navigate in
the latent space seems an important and desired feature for generative models in
creative applications.


\section{Discussion and Conclusion}
\label{sec:conclusion}
In this paper, we introduced a new regularization function on the latent space
of a VAE. This geodesic latent space regularization aims at binding a
displacement in some directions of the latent space to a qualitative change of
the attributes of the decoded sequences. We demonstrated its efficiency on a
music generation task by providing a way to generate variations of a given
melody in a prescribed way.

Our experiments shows that adding this regularization allows interpolations in
the latent space to be more meaningful, gives a notion of geodesic distance to
the latent space and provides latent space variables with less correlation
between its regularized and non-regularized coordinates.

Future work will aim at generalizing this regularization to variational
autoencoders with multiple stochastic layers. It could indeed be a way to tackle
the issue of inactive units in the lower stochastic layers as noted in
\cite{chen2016variational,maaloe2017semi}, by forcing these lower layers
to account for high-level attributes.

Our regularization scheme is general and can be applied to the most recent
generalizations of variational autoencoders which introduce generative
adversarial training in order to obtain better approximations of the posterior
distributions \cite{makhzani2015adversarial,2017arXiv170604223J} or in
order to obtain a better similarity metric \cite{larsen2015autoencoding}.

We believe that applying this regularization to conditional VAEs
\cite{NIPS2015_5775,2015arXiv151200570Y} opens up new ways to
devise interactive applications in a variety of content generation tasks.

\section{Acknowledgments}
\label{sec:acknowledgements}
First author is founded by a PhD scholarship of \'{E}cole Polytechnique (AMX).

\bibliographystyle{abbrv}
\bibliography{glv}

\appendix
\section{Implementation Details}
\label{sec:details}
We report the specifications for the model used in Sect.~\ref{sec:exp}.  All
RNNs are implemented as 2-layer stacked
LSTMs \cite{hochreiter1997long,mikolov2014learning} with 512 units per layer and
dropout between layers; we do not find necessary to use more recent
regularizations when specifying the non stochastic part model like Zoneout
\cite{krueger2016zoneout} or Recurrent batch normalization
\cite{cooijmans2016recurrent}.
We choose as the probability distribution $r_1$ on the partial gradient norm a
normal distribution with parameters $\mathcal{N}(2, 0.1)$.

 We find out that the use of KL annealing was crucial in the training
 procedure. In order not to let the geodesic latent space regularization too
 important at the early stages of training, we also introduce an annealing
 coefficient for this
 regularization. This means we are maximizing the following regularized ELBO
 \begin{equation}
   \label{eq:klannealing}
   \mathbf{E}_{q_\phi(z|x)} \left[ \log p(x |z) + \beta \, \mathcal{R}_{\textrm{geo}}(z;
     \{g_k\}, \theta)\right]
   - \beta \, D_{KL}(q(z|x) || p(z))
 \end{equation}
 with $\beta$ slowly varying from $0$ to $1$. We also observe the necessity of
 early stopping which prevents from overfitting.

\section{Choice of the regularization parameters}
\label{sec:regpar}

We highlight in this section the importance of the choice of the regularization
distributions $r_k$ on the partial gradients (Eq.~\ref{eq:glsrk}). Typical choices
for $r_k$ seem to be univariate normal distributions $\mathcal{N}(\mu, \sigma^2)$
of mean $\mu > 0$ and standard deviation $\sigma$.

\begin{figure}[h]
  \centering
    \includegraphics[scale=0.3]{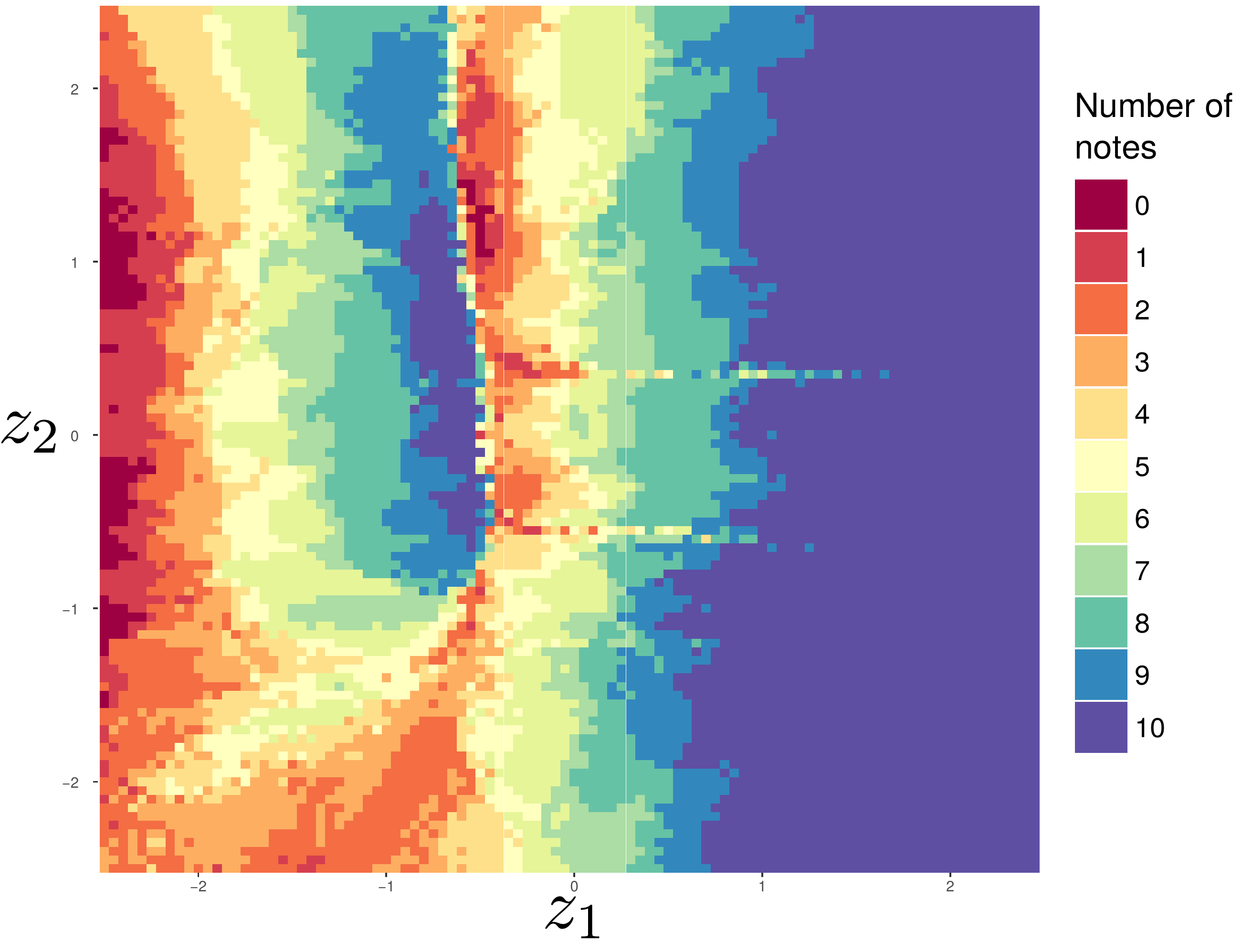}
\caption{Same plot as in Fig.~\ref{fig:numNotesReg} but with $r_1 = \mathcal{N}(5, 1)$.}
  \label{fig:numNotesTwoPhases}
\end{figure}

In practice, this distribution's mean value must be adapted to the values of the attribute
functions $g_k$ over the dataset. Its variance must be small enough so that the
distribution 
effectively regularizes the model but high enough so that the model's
performance is not drastically reduced and training still possible.
Fig.~\ref{fig:numNotesTwoPhases} displays the same figure as in
Fig.~\ref{fig:numNotesReg}, but with a regularization distribution $r_1$ being a
normal distribution $\mathcal{N}(5, 1)$. We see that imposing too big an
increase to fit in the prior's range can cause unsuspected effects: we see here a
clear separation between two phases.

What is the best choice for the $r_k$ distributions depending on the values of the $g_k$ functions
over the dataset is an open question we would like to address in future works.

\end{document}